# An Incremental Explanation of Inference in Bayesian Networks for Increasing Model Trustworthiness and Supporting Clinical Decision Making

Evangelia Kyrimi[1], Somayyeh Mossadegh[2], Nigel Tai[3, 4], William Marsh[1]

[1]School of Electronic Engineering & Computer Science, Queen Mary University of London, Mile End Road, London E1 4NS, UK
[2]Centre for Trauma Sciences, Blizard Institute, Queen Mary University of London, The Blizard Building, 4 Newark Street, London E1 2AT, UK
[3]Royal London Hospital, Barts and The London NHS Trust, London, E11BB
[4]Academic Department of Military Surgery and Trauma, Royal Centre for Defence Medicine, Birmingham Research Park, Vincent Drive, Birmingham B152SQ

Abstract

Various AI models are increasingly being considered as part of clinical decision-support tools. However, the trustworthiness of such models is rarely considered. Clinicians are more likely to use a model if they can understand and trust its predictions. Key to this is if its underlying reasoning can be explained. A Bayesian network (BN) model has the advantage that it is not a black-box and its reasoning can be explained. In this paper, we propose an incremental explanation of inference that can be applied to 'hybrid' BNs, i.e. those that contain both discrete and continuous nodes. The key questions that we answer are: (1) which important evidence supports or contradicts the prediction, and (2) through which intermediate variables does the information flow. The explanation is illustrated using a real clinical case study. A small evaluation study is also conducted.

## 1. Introduction

Many clinical decision support (CDS) models have been developed in medicine [1]. However, very few of them have been actually used in practice to support decision making [2], [3]. Sometimes it is assumed that an accurate prediction is enough for making a CDS model useful, but this neglects the importance of trust [4], [5]. A user, who does not understand or trust a model, will not accept its advice [2], [5]. The lack of trust may be due to the difficulty of understanding how a prediction is inferred from the given data. As Aristotle wrote 'we do not have knowledge of a thing until we have grasped its why, that is to say, its explanation'. Hence, explaining a model's reasoning, its inference, could increase trustworthiness.

This paper considers Bayesian Networks (BNs): directed acyclic graphs, showing causal or influential relationships between random variables. These variables can be discrete or continuous, and a BN with both is called `hybrid'. The uncertain relationships between connected variables are expressed using conditional probabilities. The strength of these relationships is captured in the Node Probability Table (NPT), used to represent the conditional probability distribution of each node in the BN given its parents. Once values for all NPTs are given, the BN is fully parameterized, and probabilistic reasoning (using Bayesian inference) can be performed. However, the reasoning process is not always easy for a user to follow [6], [7], [8].

In contrast to many CDS models, a BN is not a black box and its reasoning can be explained [6], [9]. Several approaches have been proposed to explain the reasoning of a BN (presented in Section 3). However, there are many situations where these methods cannot be applied. First, most of the methods are applied to BNs that include only discrete variables. Some of them are even restricted to binary variables only. However, most of the medical BNs include continuous nodes as well. In addition, most of them try to find the best explanation that can be time-consuming, especially for large BNs, which are common in medical applications. Finally, in some methods, user input is required in different stages of the explanation. This can be problematic, especially in situations where there is time pressure.

In this paper, we propose a practical method of explaining the reasoning in a BN, so that a user can understand how a prediction is generated. The method is an extension of a previous conference paper published by the authors [10]. Our proposed method can be used in hybrid networks that have both continuous and discrete nodes and requires no user input. In addition, we simplify the process of identifying the most important evidence and chains of reasoning, so we can rapidly produce a good and concise explanation, but not necessarily the most complete one. In fact, our method produces an incremental explanation that has three successive levels of detail. The key questions that we answer are: (1) which important evidence supports or contradicts the prediction, and (2) through which intermediate variables does the information flow. A clinical case study on predicting coagulopathy in the Emergency Department (ED) is used to illustrate our explanation. An evaluation study of the impact that the explanation has on clinicians' trust is also presented using the same case study.

This paper is organised as follows: Section 2 presents necessary background material, explaining how we are using the idea of explanation to increase a user's trust in a clinical decision support model implemented using a BN. In particular, we contrast our objectives with the wider literature on 'explanation'. Section 3 presents related works that shares our objectives. The proposed method is presented in Section 4. The verbal output of the explanation is illustrated using a real scenario in Section 5. Section 6 shows a small evaluation study. Discussion and conclusion are presented in Section 7 and 8, respectively.

## 2. Background: Conditional Independence, Markov Blanket and Explanation in Bayesian Networks

This section introduces necessary preliminaries for understanding the proposed method. First, a brief description of conditional independences in BNs is presented. Second, the Markov Blanket (MB) of a variable in a BN is defined. Finally, the different objectives of an explanation are described, noting how these apply in the context of BNs.

### 2.1 Conditional Independence

Two variables *A* and *B* are independent, usually written as $A \perp B$, if and only if their joint probability equals the product of their probabilities, that is $P(A,B) = P(A)P(B)$. Equivalently, $A \perp B$ if $P(A|B) = P(A)$. Two variables *A* and *B* are conditionally independent given a third observed variable *C* if $P(A,B|C) = P(A|C)P(B|C)$. In a probability distribution represented by a BN, some conditional independence relationships can be determined using the graphical criteria of d-separation [11]. Given a set of observed variables, variables that are not d-separated are said to be d-connected.

Suppose that we have three variables *A*, *B* and *C* and we want to know whether the variables *A* and *C* are d-separated given the variable *B*. There are three types of connection as shown in Figure 1:
- **Serial connection** (Figure 1a): *A* and *C* are d-separated, given that *B* is observed.
- **Diverging connection** (Figure 1b): *A* and *C* are d-separated given that *B* is observed.
- **Converging connection** (Figure 1c): *A* and C are d-separated only if the variable B or any of its descendants are not observed.

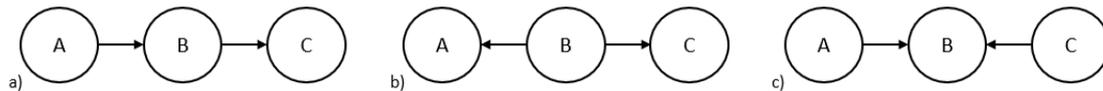

*Figure 1. (a) Serial connection. (b) Diverging connection. (c) Converging connection*

When variables *A* and *C* are d-separated, then the information from variable *A* does not flow to variable *C* and vice-versa.

## 2.2 Markov Blanket

The term Markov Blanket (MB) applied to BNs was first introduced by Judea Pearl in 1988 [11]. The MB of a variable in a graphical model contains all the variables that can d-separate it from the rest of the network. In other words, the MB of a variable is the only knowledge needed to predict that variable and its children.

In BNs the MB of a variable *A*, denoted as *MB(A)*, consists of its parents, children and children's other parents as shown in Figure 2. The values of the parents and children of a variable evidently give information about that variable. However, its children's other parents also must be included, because they can be used to explain away the variable in question.

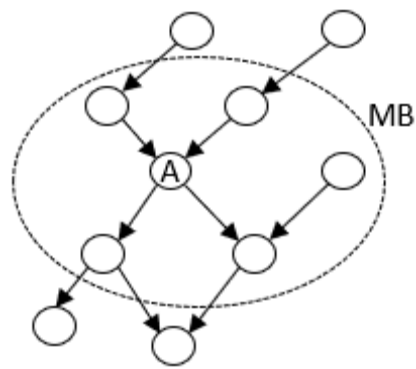

*Figure 2. The Markov Blanket of variable A in a Bayesian Network*

## 2.3 Explanation in Bayesian Networks

The concept of 'explanation' has plagued philosophers. One stream of work [12], [13], [14], [15], [16] aims to find a hypothesis that best (in some specified sense) explains some observed facts. However, this 'inference of the best explanation' (IBE) is not our objective here. In this paper, we follow the classification of the focus of explanation methods in the context of BNs as proposed by Lacave and Diez [6]. According to them an explanation in the context of BNs can be focussed on any of the following objectives, the first of which corresponds to IBE:

- **Explanation of the evidence**: '*it consists of determining which values of the unobserved variables justify the available evidence. This process is usually called*

*abduction, and it is based on the (usually implicit) assumption that there is a causal model*' [6].
- **Explanation of the model**: '*it consists of displaying (verbally, graphically or in frames) the information contained in the knowledge base*' [6]. In other words, we explain how the structure and parameters of the model relate to domain knowledge [17].
- **Explanation of reasoning**: '*it may provide three kinds of justification; (1) the results obtained by the system and the reasoning process that produced them, (2) the results not obtained by the system, despite the user's expectations, (3) hypothetical reasoning*' [6].

All the above types of explanation could be useful as part of decision support. However, as our aim is to increase the user's trust in the model's prediction, we wish to explain the model's reasoning. This is the third in the list of possible focusses of an explanation in BNs. The form of such an explanation can be illustrated by the following scenario:

> *A doctor uses a BN that predicts the likelihood of coagulopathy[1] in traumatically injured patients. He enters the evidence and the model predicts that the patient is 8.7 times more likely to become coagulopathic than an average trauma patient. When asked to explain, the system informs him that, despite the positive effects of the absence of a long bone and pelvic fracture and a negative FAST scan[2], the likelihood of coagulopathy increased because of haemothorax (blood in the chest) from rib fractures and lung injury, the high energy of the injury, a base excess[3] of -14, a Glasgow Coma Scale[4] (GCS) of 4 and the administration of more than 500ml of fluids. In complicated cases, just explaining the significant positive and negative evidence may not be sufficient. The system can further explain that the evidence affected the prediction of coagulopathy through the unobserved variables tissue injury and tissue perfusion.*

This example shows the basic components of an explanation. First, the explanation has a statement that needs to be explained, we will refer to it as *target*. In this example, the target is the reported high odds that the patient is coagulopathic. Then evidence that supports or contradicts the reported statement is presented. Finally, additional information that is not directly observed but plays an important role in clinical reasoning is described.

### 3. Related work on Explanation of Reasoning in a BN

Several methods of explaining the reasoning in a BN have been proposed. Most of these methods include the following elements: i) measuring the **impact** of the evidence variables on the target, ii) determining a **threshold** for selecting the variables that should be included in the explanation, iii) distinguishing between supporting and **conflicting** evidence and iv) explaining the flow of information from evidence variables to the target, described as '**chains of reasoning**'. Our proposed explanation method is also based on these elements as it places the evidence at the core of the explanation, which is appropriate in a clinical decision support context. Several variants of this approach have been proposed. For instance, Yap et al. [17] shift the explanation entirely to the MB; this idea has been incorporated in our approach, but we use it alongside the evidence variables.

---

[1] Coagulopathy is a bleeding disorder, in which the blood's ability to coagulate (form clots) is impaired.
[2] FAST scan is a diagnostic test for internal intra-abdominal free fluid.
[3] Base excess is used as a sign of respiratory problems.
[4] GCS is a sign of consciousness.

Other approaches for explaining the reasoning in BNs are quite different. In [18], Shih et al. create an explanation by analysing a classifier's 'decision function', which can be understood as an exhaustive analysis of the cases giving a positive classification. A disadvantage of this approach for our purposes is that the explanation depends on the operating point of the classifier, whereas the output of a decision support system may vary smoothly with the posterior probabilities (e.g. using traffic lights) rather than using a single operating point.

Another approach for explaining the reasoning in BNs is based on argumentation theory, widely used in legal cases. Arguments are extracting from a BN though an argument 'support graph' [18], [19], [20], [21] from which an argument graph is extracted depending on the evidence. Although this representation has a history of application to the law, as far as we are aware it has not been widely applied in medicine.

In the remainder of this section, explanation methods that cover some or all the elements introduced above are reviewed in detail.

## 3.1 Evidence impact

Not all the evidence has equal impact on the target variable. Measuring the impact involves assessing the change in the probability distribution of the target produced by the evidence; there are different distributions that can be compared and different measures to do that. The INSITE method, proposed by Suermondt [22], uses the KL divergence between the posterior of the target with all the evidence and the posterior of the target when each evidence (one-way analysis) or a subset of evidence (multi-way analysis) has been temporarily removed. Exact multiway analysis for the best subset of evidence is time consuming as is exponential in the number of evidence variables. In addition, the KL divergence is ill-defined when the denominator is 0. Chajewska and Draper address the computational complexity with more flexible requirements for the size of the explanation set and the significance of the impact that each evidence variable has on the target [23]. They also point out that the prior probability of the target needs to be considered. BANTER [24] measures the difference between the prior and the posterior of the target for each evidence variable on its own. However, this simplification can be misleading sometimes as it neglects the rest of the available evidence variables. Madigan *et al.* assess the impact using Good's weights of evidence [25], evaluated incrementally as the user instantiates each evidence variable; a binary target is assumed, and the calculated weights depend on the order the evidence is entered [26].

## 3.2 Significance Threshold

The explanation should only include the evidence variables with the greatest impact. Many ways have been proposed to find an appropriate impact threshold. A simple approach is for the end user to choose a threshold [23]. However, even if the end-user has the domain knowledge needed, it is hard for him to express this in terms of the range of the distance measurement. Alternatively, a fixed threshold is chosen by the model builder [24] or the impact of all the evidence variables is presented, from the largest to the smallest, without a threshold [27]. This can make the explanation very complex when there are many evidence variables. INSITE proposes an indirect way of specifying a significance threshold. Instead of choosing an appropriate threshold for the distance measurement, the user specifies an 'indifference' range for the posterior of the target; changes outside this range are significant and the corresponding threshold can be calculated. This approach combines the users' domain knowledge, given as the range of indifference on the probability, and the characteristics of the distance measure. However, this range may need to be changed for each query and it is still not easy for the end user to do this, especially when the target variable is continuous, or the decision tool is being used under time pressure.

## 3.3 Conflict Analysis

We also want to know whether each evidence variable supports or conflicts with the overall change predicted by the model. INSITE introduced the idea of conflict analysis in an explanation, looking at whether removing an evidence variable shifts the posterior of the target in the same direction as the change from the posterior with all the evidence to the prior when all the evidence is removed. However, this analysis is limited to binary variables. For non-binary variables mixed effects can occur, where the change for some states supports, and for other states conflicts with, the overall change. Madigan's use of the weight of evidence distinguishes between positive and negative effects, but it may depend on the order evidence is entered.

## 3.4 Chains of Reasoning

Evidence variables may be connected to the target by other non-observed variables in a 'chain of reasoning'. Choosing which of these variables to include in the explanation is difficult as there can be many such chains. INSITE generates a set of directed chains from each significant evidence variable to the target. Then it calculates the difference between the prior and posterior marginal distributions for each variable included in the chain. Based on a screening rule, a chain is eliminated if there is at least one variable for which there is no substantial difference (the difference is less than the threshold of significance as explained in Section 3.2). The elimination is because the variable acts as though it blocks the evidence transmission through the chain. In the remaining chains, additional screening is performed by carefully removing arcs. By analysing the difference between the prior distributions of the target with and without the arc in comparison with the changes after the evidence transmission, the effect of the removed arc - and of the chains that include the arc - on the transmission from the evidence to the target was derived. BANTER selects the chains with the highest strengths and the minimum length (among chains with the same strength) by measuring the impact of every variable in the chain. The strength of the chains is given by the minimum impact of any of the variables in the chain. Madigan et *al*. screen the evidence chains by looking at the weight of evidence of every variable in a chain of reasoning. The weight of evidence for each variable relates to the ratio between the weights of the incoming and outgoing evidence. However, they only consider networks with a tree structure, which have only a single path from an evidence variable to the target. Leersum tries to find a non-empty set of intermediate variables that summarizes all the information between the evidence and the target [28]. He looks at the weight of the edges using a Maximum-flow-minimum-cut theorem and then considers only the variables that are connected with the edges of the minimum cut, which is the minimum set of edges that makes the graph disconnected.

# 4 Generating an Incremental Explanation of Reasoning

This section presents the algorithm for producing an incremental explanation of reasoning in BNs. First, an overview of the overall algorithm and the levels of the explanation are introduced. Then the algorithm for each level of the explanation is described in detail.

## 4.1 Overview

In every scenario, we have a target variable $T$ for which we compute the posterior probability distribution given a set of observed evidence $E$. There is also a set of unobserved intermediate evidence variables $I$ that are part of the flow of reasoning from $E$ to $T$. The variables that are included in the explanation are called the explanatory variables $X$. The set $X$ consists of a set of significant evidence $E_{sig}$, which are d-connected to $T$ and have a significant impact on it (explained in detail in Section 4.2), and a set of important intermediate variables $I_{sig}$ that are unobserved (i.e. not evidence variables) and act as a middle step in the flow of information from $E_{sig}$ to $T$ (explained in detail in Section

4.3 and 4.4). The different sets of variables are shown in Figure 3. In case more than one *T* variables are available. Then an explanation for each of them is produced separately.

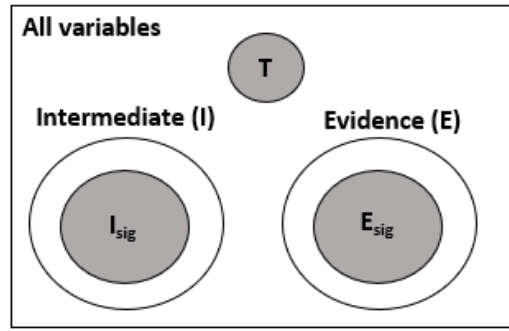

*Figure 3. Set of variables in the explanation of reasoning, with explanatory variables $X = I_{sig} \cup E_{sig}$*

As illustrated in Figure 4, the proposed explanation has three levels of increasing detail:

1. *Level 1 – Significant evidence variables*: The first level lists the significant evidence variables $E_{sig}$, ordered by their impact on *T*. The variables presented in this level are grouped into two clusters based on whether they support or conflict with the effect of the combined evidence.
2. *Level 2 – Information flow*: The second level identifies the intermediate variables $I_{sig}$ through which the information from $E_{sig}$ to *T* flows and it shows how the evidence has changed the probability distribution of $I_{sig}$.
3. *Level 3 – Significant evidence impact on the intermediate variables*: The third level describes the impact of each member of $E_{sig}$ on each of the intermediate variables $I_{sig}$ supporting or conflicting with the combined effect.

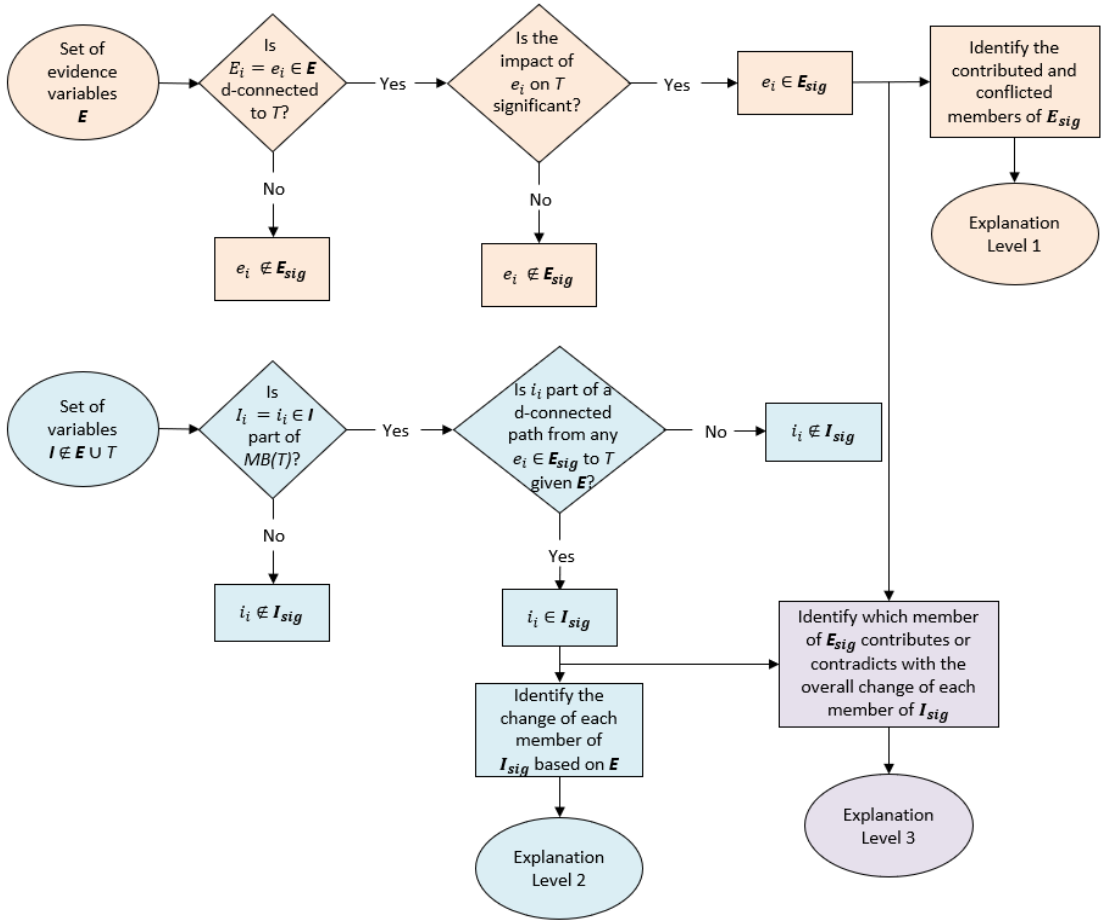

*Figure 4. The process of the proposed explanation of reasoning*

## 4.2 Level 1: Significant Evidence Variables

In the first level of the explanation, we try to answer the question 'How does each evidence affect the target?'. Answering that question requires first a measure of impact, then a threshold of significance and finally an analysis of whether each evidence supports or conflicts with the overall change.

### 4.2.1 Evidence Impact

Following INSITE, the impact of an evidence variable $E_i=e_i$ relates to the distance between the posterior probability with all the evidence $P(T|E)$ and the marginal posterior probability when $e_i$ is excluded from the set of evidence $P(T|E\backslash e_i)$, which we denote as:

$$Im_E(e_i) \triangleq D(P(T|\boldsymbol{E})||P(T|\boldsymbol{E}\backslash e_i)) \qquad (1)$$

In case, we are interested about the distance between the posterior probability with all the evidence $P(T|E)$ and the prior probability when the whole set of evidence is excluded $P(T) = P(T|E\backslash E)$, then equation (1) is defined as:

$$Im_E(\boldsymbol{E}) \triangleq D(P(T|\boldsymbol{E})||P(T)) \qquad (2)$$

INSITE uses the KL divergence as the distance metric. However, it is not always well defined. In our proposed method, the difference between the two distributions is measured using Hellinger distance ($D_H$). For instance, given two discrete distributions $P: \{p_1, \ldots p_n\}$ and $Q: \{q_1, \ldots q_n\}$, Hellinger distance is defined as:

$$D_H(P,Q) = \frac{1}{\sqrt{2}} \sqrt{\sum_{i=1}^{n} (\sqrt{p_i} - \sqrt{q_i})^2} \qquad (3)$$

In case *P* and *Q* are continuous distributions, the square of the Hellinger distance between *P* and *Q* is defined as:

$$D_{H^2}(P,Q) = \frac{1}{2} \int \left(\sqrt{dP} - \sqrt{dQ}\right)^2 \qquad (4)$$

Where $\sqrt{dP}$ and $\sqrt{dQ}$ denote the square root of the densities. For instance, the Hellinger distance between the posterior and the prior distributions presented in Figure 5 is $D_H(P(T|E = False)||P(T)) = 0.42$. The distributions presented in Figure 5 are generated following the dynamic discretization algorithm [29] available in AgenaRisk [30]. However, the same approach can be used in situations where the histogram of a distribution can be constructed from whatever underlying representation of the probability distribution is been used by the inference system.

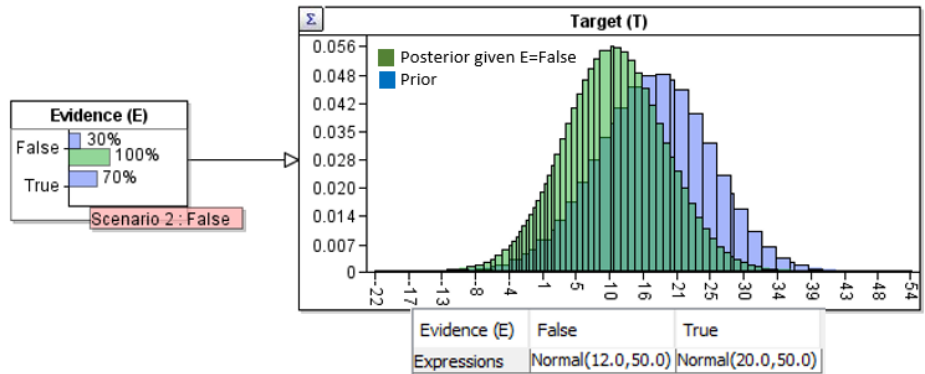

*Figure 5. A two-node BN example, in which the posterior distribution of the target given E=False is superimposed against the prior distribution.*

Hellinger distance is symmetric, non-negative and it satisfies the triangle inequality. It ranges between 0 and 1. This distance metric was used for various reasons; (i) it can be calculated for both discrete and continuous distributions, (ii) it is u-shaped, (iii) it is always well defined. Having a distance metric that can be applied to both discrete and continuous distributions is fundamental, as nowadays more networks are 'hybrid'. A u-shape metric gives a greater penalty to the distance from 0.9 to 0.91 than from 0.5 to 0.51. This is appropriate since a probability near either 0 or 1 represents near certainty. Finally, the distance metric should be defined for all the values of the two compared distributions. Other distance metrics that have these properties can be used as well.

### 4.2.2 Significance Threshold

The proposed approach for specifying the *threshold of significance* is an extension of the INSITE method. The challenge, introduced in Section 3.2, is to find an approach that applies in all cases. In the INSITE method, a sequence of definitions is made to achieve a parameter that defines significance and can also be interpreted by user. The user is then asked to select the value of this parameter. We wish to avoid this user input step, so we follow the same sequence of definitions but automate the final step using a simple heuristic.

The first definition is a threshold $\theta$ for the impact of an evidence variable $e_i$. Specifically, $\theta$ is the minimum impact for evidence to be considered significant:

$$e_i \in E_{sig} \; iff \; Im_E(e_i) \geq \theta \qquad (5)$$

Since it is not easy to set a single threshold $\theta$ for all cases, it is specified indirectly by first defining a marginal posterior probability $G$, which lies within the direction of change from $P(T|E)$ to $P(T)$, as shown in Figure 6, such as $\theta \triangleq D_H(P(T|E)||G)$. $G$ is in turn determined by a parameter $\alpha$, defining $G$ as a proportion of the difference between $P(T|E)$ and $P(T)$:

$$G \triangleq P(T|E) - \alpha(P(T|E) - P(T)) \qquad (6)$$

In INSITE, the user chooses $\alpha$. To avoid the user input and make the generation of the explanation fully automatic, a predefined set of decreasing $\alpha$: $\{\alpha_1 \ldots \alpha_n\}$ is introduced. Starting with the largest $\alpha_i$, each smaller value is tested until at least half of the evidence variables are included in $E_{sig}$.

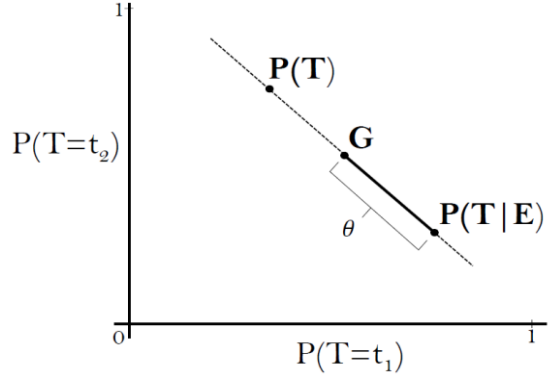

Figure 6. Threshold of significance for a binary target T (based on Suermondt 1992)

To illustrate this, imagine that we have a BN with a binary target $T$: $\{t_1, t_2\}$ and a set $E$ of 6 evidence variables $E$: $\{E_1=e_1 \ldots E_6=e_6\}$. The prior, marginal posterior and posterior probabilities are: $P(t_1) = 0.097$; $P(t_1|E\backslash e_1) = 0.19$; $P(t_1|E\backslash e_2) = 0.15$; $P(t_1|E\backslash e_3) = 0.27$; $P(t_1|E\backslash e_4) = 0.11$; $P(t_1|E\backslash e_5) = 0.21$; $P(t_1|E\backslash e_6) = 0.26$; $P(t_1|E) = 0.2$. Using the Hellinger distance, as defined in equation (3), the impact of each evidence (defined in equation (1)) is calculated. Then for a predefined set $\alpha$: $\{0.5, 0.45, 0.4, 0.35, 0.3, 0.25, 0.2, 0.15, 0.1, 0.05, 0.01, 0.005, 0.001\}$ a posterior distribution $G$ is calculated following the rule defined in equation (6). For each G, a threshold $\theta$ is calculated, such as $\theta \triangleq D_H(P(T|E)||G)$. As shown in Figure 7, we have $E_{sig}$: $\{e_4, e_3, e_6\}$ based on $\alpha = 0.5$ and $\theta = 0.048$. If, instead, the KL divergence is used, the set $E_{sig}$ and $\alpha$ remain the same but the threshold of significance $\theta$ changes to 0.0042. Although in this example the same set of significant evidence was found in, the example illustrates the difficulty of defining a fixed $\theta$. In addition, $\theta$ also depends on the evidence as the same distance metric and the same $\alpha$ can lead to very different $\theta$ values in different scenarios.

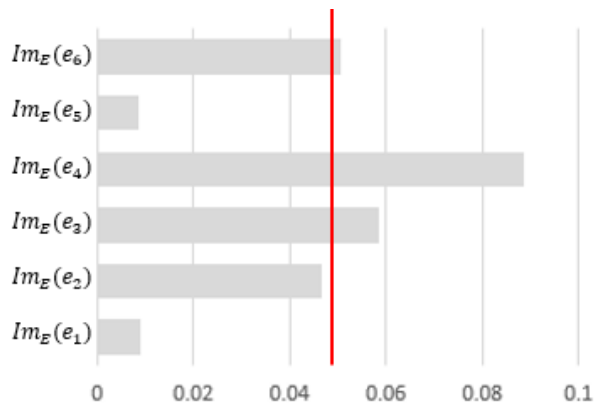

*Figure 7. Impact of each evidence variable. The vertical red line shows the threshold $\theta = 0.48$; evidence variables with an impact exceeding this threshold are considered significant.*

### 4.2.3 Conflict Analysis

Having identified the set of significant evidence variables $E_{sig}$, we next examine whether each evidence variable works in the same way in creating the overall change of *T*. This is known as 'conflict analysis' and we extend INSITE's method to work for variables with more than two states. When we perform a conflict analysis we compare (i) the direction of the change and (ii) the impact on the target when each evidence variable is removed with the impact when all the evidence variables are removed. The direction of change can be assessed using the difference $\Delta_{t_i}(e_i)$ for every state $t_i$ of *T* defined as:

$$\Delta_{t_i}(e_i) = P(t_i|E) - P(t_i|E \setminus e_i)) \tag{7}$$

For each state $t_i$, the difference $\Delta_{t_i}(e_i)$ is compared to the difference $\Delta_{t_i}(E) = P(t_i|E) - P(t_i)$. If both differences have the same sign for each $t_i$, then the direction of the change is consistent (Equations 8 and 10). If for each $t_i$ the sign of these distances is the opposite, then the direction is conflicting (Equations 9 and 11). Finally, when the sign of these differences is not the same or the opposite for each $t_i$, then the direction is mixed (Equation 12). Imagine, for instance that we have the target variable *B*: {$b_1$, $b_2$, $b_3$} as shown in Figure 8. For the state $b_1$, both differences have a consistent positive sign. Similarly, the differences for state $b_3$ have a consistent negative sign. However, for state $b_2$ the direction of change when we remove one evidence and when we remove all the evidence, is positive and negative, respectively. As a result, the direction of change is mixed. This gives the following definitions for the direction of change:

$$d_{\text{consistent}}(e_i, t_i) = \text{sgn}(\Delta_{t_i}(e_i)) = \text{sgn}(\Delta_{t_i}(E)) \tag{8}$$

$$d_{\text{conflicting}}(e_i, t_{i\ i}) = \text{sgn}(\Delta_{t_i}(e_i)) \neq \text{sgn}\left(\Delta_{t_i}(E)\right) \tag{9}$$

$$D_{\text{consistent}}(e_i) = \forall t_i . d_{\text{consistent}}(e_i, t_{i_i}) \tag{10}$$

$$D_{\text{conflicting}}(e_i) = \forall t_i . d_{\text{conflicting}}(e_i, t_i) \tag{11}$$

$$D_{\text{mixed}}(e_i) = \neg D_{\text{consistent}}(e_i) \land \neg D_{\text{conflicting}}(e_i) \tag{12}$$

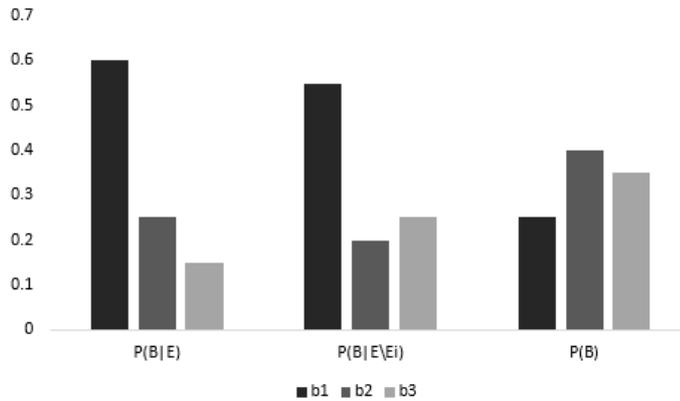

*Figure 8. Example of a mixed direction of change*

The magnitude of the impact also needs to be considered. If all the evidence variables are working together, and the direction is consistent, the impact when one variable is unobserved is expected to be less than the impact when all the evidence variables are unobserved, such as Im$_E(e_i)$ ≤ Im$_E(E)$. However, it is also possible that removing the evidence $e_i$ can lead to a greater impact than Im$_E(E)$, even though the direction is consistent. This suggests that $e_i$ 'dominates' the remaining evidence. In case the direction of change is mixed, we can assess which effect is more significant by comparing the impact of the consistent and the conflicting part. Suppose we have several states of T which $t_i \in d_{consistent}(e_i, t_i)$. The impact of the consistent part is defined as $\text{Im}_E(e_i)_{t_i} \triangleq D(P(t_i|E)||P(t_i|E\setminus e_i))$. Table 1 summarises the conflict categories.

| Conflict Category | Direction | Impact |
|---|---|---|
| Dominant | $D_{consistent}$ | $\text{Im}_E(e_i) > \text{Im}_E(E)$ |
| Consistent | $D_{consistent}$ | $\text{Im}_E(e_i) \leq \text{Im}_E(E)$ |
| Conflicting | $D_{conflicting}$ | n/a |
| Mixed consistent | $D_{mixed}$ | $\text{Im}_E(e_i)_t \mid t \in d_{consistent}(e_i, t) > \text{Im}_E(e_i)_t \mid t \in d_{conflicting}(e_i, t)$ |
| Mixed conflicting | $D_{mixed}$ | $\text{Im}_E(e_i)_t \mid t \in d_{consistent}(e_i, t) \leq \text{Im}_E(e_i)_t \mid t \in d_{conflicting}(e_i, t)$ |

Table 1. Summary of the conflict analysis categories

### 4.3 Level 2: Information Flow

The second level of the explanation uses a simple approach to present the flow of reasoning from each member of $E_{sig}$ to T. First, a set of intermediate variables $I_{sig}$ is determined (Figure 9). The MB(T) is chosen as the potential set of $I_{sig}$. From the MB variables we include in $I_{sig}$ only those that are unobserved (Figure 9a, 9b) and part of a d-connected path from $E_{sig}$ to T, given the evidence variables E (Figure 9c). In the second level of the explanation, the change in the uncertainty of each member of $I_{sig}$ is also shown. If the set $I_{sig}$ is empty (e.g. all the MB variables are observed), the explanation stops at the first level.

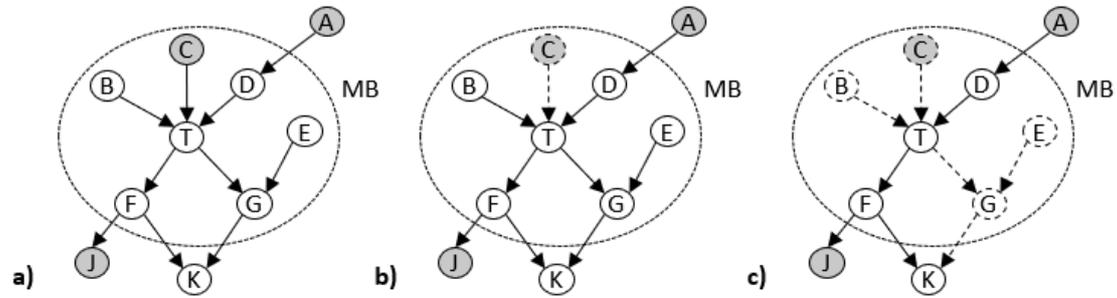

Figure 9. Process of finding the set of intermediate variables $I_{sig}$. a) The variables A, C and J are observed. b) The variable C is observed, so it is excluded from the set of $I_{sig}$. c) The variables B, E and G are not part of a d-connected part from the evidence to T, so they are excluded from the set $I_{sig}$, which is {D, F}

### 4.4 Level 3: Significant Evidence Impact on the Intermediate Variables

The final level of the explanation repeats some parts of the analysis of level 1 on the intermediate variables of level 2. For simplicity and consistency, we do not reassess the set of $E_{sig}$ for each $I_{sig}$. Instead, for each variable in $I_{sig}$, we first determine the subset of $E_{sig}$ that are d-connected to them given E, and we carry out the conflict analysis as described before.

## 5 Case Study

The output of our explanation method is illustrated using a clinical case study on acute traumatic coagulopathy, a bleeding disorder in which the blood's ability to clot is impaired.

## 5.1 Detecting Coagulopathy

The aim of the developed BN was to predict coagulopathy in the first 10 minutes of hospital care [31]. All the variables that may be observed within 10 minutes are shown in purple (Figure 10). The target variable, COAGULOPATHY, is shown in red. There are 11 variables in the MB of the target. The variables PREHOSP and AGE are observed, so they are excluded from the set of $I_{sig}$. In addition, the variables ROTEMA30 to APTTr (see top right) are not part of the flow of reasoning, while DEATH and HEAD are also not part of a d-connected path between the target and any of the evidence variables; this results in two intermediate variables: ISS (tissue Injury Severity Score) and PERFUSION (oxygen delivered to the tissues of the body).

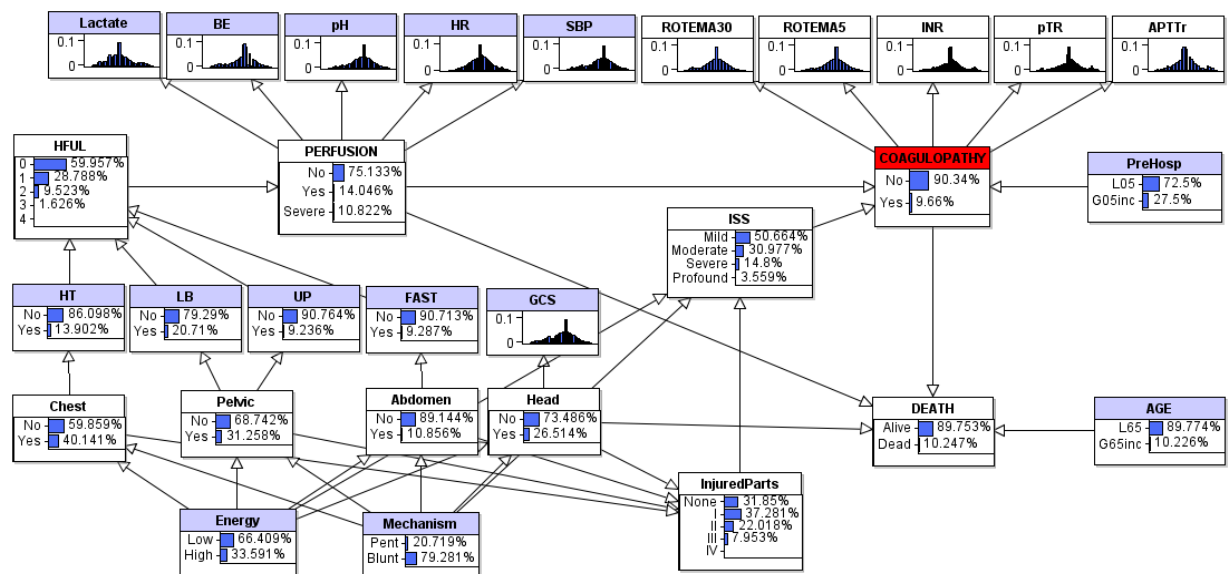

*Figure 10. A BN model that predicts coagulopathy within 10 minutes of hospital care*

## 5.2 Verbal Output

The output of our algorithm is verbal and consists of three main parts: (i) numerical data, (ii) fixed text and (iii) dynamic text. Numerical data are presented using numbers. Fixed text consists of standard phrases that can be repeated in different scenarios and are presented with small letters. Dynamic text is different in every scenario and is shown with capital letters. At the beginning, the prediction of the target is presented. By default, the state with the highest probability is presented. If the user wants to know the output of a specific state, he can configure it. In this case study, clinicians were interested only in the likelihood of having coagulopathy. Then supporting and conflicting significant evidence variables are presented in a decreasing order based on their significance. We can have up to four groups of significant evidence: (i) consistent and dominant evidence, (ii) conflicting evidence, (ii) mixed consisting evidence and (iv) mixed conflicting evidence (see Table 1)

**The likelihood of COAGULOPATHY = YES is 11%.**
**This patient has a 14% INCREASE in risk of becoming coagulopathic than an average trauma call patient.**

**Factors that support the INCREASED risk of COAGULOPATHY = YES (strongest to least):**
- PREHOSPITAL FLUIDS ≥ 500mls (Very important)
- GCS = 5                    (Very important)
- HAEMOTHORAX = YES          (Very important)
- ENERGY OF INJURY = HIGH

**Factors that do not support the INCREASED risk of COAGULOPATHY = YES (strongest to least):**

- **SYSTOLIC BLOOD PRESSURE = 168**
- **LONG BONE FRACTURE = NO**
- **LACTATE = 0.9**

Coagulopathy is a binary variable, so no mixed effects are present. The significant evidence variables are described as 'supporting' or 'not supporting'. Supporting evidence is the consistent and dominant (distinguished by the phrase 'Very important') variables as described in Section 4.2.3. Non-supporting variables are those classed as conflicting in Section 4.2.3. In this scenario, where the patient has an increased risk of becoming coagulopathic, compared with the average trauma patient (prior), the supporting evidence increases the risk, while the non-supporting evidence decreases it. Then the intermediate variables and their change are presented.

**Important elements for predicting COAGULOPATHY are:**
1. PERFUSION: The likelihood of PERFUSION = NORMAL is 95%

This patient has a 26% INCREASE in risk of having PERFUSION = NORMAL than an average trauma call patient.

2. ISS: The likelihood of ISS = SEVERE is 49%

This patient has a 230% INCREASE in risk of having ISS = SEVERE than an average trauma call patient.

Level 2 shows how the intermediate variables $I_{sig}$ have been updated by the evidence. The likelihood of the state with the highest probability is presented. Again, the output of a specific state can be configured if needed. In the last level of the explanation we present the effect that the d-connected significant evidence variables have on each intermediate variable.

| Factors that support the INCREASED risk of PERFUSION = NORMAL:<br>• SYSTOLIC BLOOD PRESSURE =168<br>• LACTATE = 0.9<br>• LONG BONE FRACTURE = NO<br>Factors that partially support the INCREASED risk of PERFUSION = NORMAL:<br>• NONE | Factors that do not support the INCREASED risk of PERFUSION = NORMAL:<br>• HAEMOTHORAX = YES<br>Factors that partially do not support the INCREASED risk of PERFUSION = NORMAL:<br>• NONE |
|---|---|
| Factors that support the INCREASED risk of ISS = SEVERE:<br>• NONE<br>Factors that partially support the INCREASED risk of ISS = SEVERE:<br>• GCS = 5<br>• HAEMOTHORAX = YES<br>• ENERGY OF INJURY = HIGH<br>• LONG BONE FRACTURE = NO | Factors that do not support the INCREASED risk of ISS = SEVERE:<br>• NONE<br>Factors that partially do not support the INCREASED risk of ISS = SEVERE:<br>• NONE |

Level 3 shows the impact that the significant evidence variables have on the intermediate variables. Since PERFUSION and ISS have more than two states, mixed effects can occur and, are presented with the terms 'partially support' and 'partially do not support'.

# 6 Evaluation

A small evaluation study was carried out with two aims: (i) compare the similarity between clinicians' reasoning and our explanation, (ii) examine the plausibility that the explanation can have a beneficial effect on clinical practice. The model presented in Section 5.1 was used as a case study.

## 6.1 Study Design

In this study, we presented 10 cases to 16 clinicians. A group of clinicians reviewed each case and selected only the cases that the model could correctly identify as true positives or true negatives. This was a before-after study organized into two parts. The first part helped us understand clinicians' reasoning and decision making. This was achieved by carrying out a baseline questionnaire for each case (160 responses). The second part assessed the potential benefit of the explanation. Each clinician completed a follow-up questionnaire; for half of the cases only the prediction of the model was presented (*prediction* cluster) and for the other half an extra explanation of the model's reasoning was given (*explanation* cluster) (Table 2).

|            | Cases       |             |
|------------|-------------|-------------|
| Clinicians | Set X       | Set Y       |
| Group A    | Prediction  | Explanation |
| Group B    | Explanation | Prediction  |

*Table 2. Each group of clinicians saw half of the cases only with the model's prediction (prediction cluster) and the other half with an extra explanation of the model's reasoning process (explanation cluster).*

To control for biases, the cases in the two randomly created sets were matched pairwise based on their complexity. The same procedure was followed for the two random groups of consultants. Clinicians were matched in each group based on their expertise. Another factor that could cause bias was the order that each case was seen. For that reason, we presented the prediction or the explanation cluster randomly to each clinician. Thus, we prevented any influence in the response with the model's explanation by their previous experience with only the prediction of the model and vice-versa.

## 6.2 Questionnaire

Each clinician completed a baseline and a follow-up questionnaire. In each questionnaire the following questions were asked:

1. What is your initial impression of this case in relation to coagulopathy?
2. Why? Rank the available information from most important to least important
3. What would you do next?

On the follow-up questionnaires we also examined the potential benefit of the extra information. The baseline questions were repeated, and we asked additional questions that rated how the extra information increased their trust of the model's prediction and how useful and clear it was.

## 6.3 Data Analysis

The primary objectives were to assess: (i) similarity between clinicians' reasoning and the explanation (*similarity*) and (ii) increase in trust in the model given an explanation (*trust*). The secondary objectives were to assess: (i) potential benefit to the clinicians' assessment and decision making given an explanation (*potential benefit*) and (ii) clarity of the explanation (*clarity*).

To assess the *similarity*, we investigated only whether the produced explanation contained all the evidence that clinicians mentioned as significant. As we restricted the set of significant evidence - evidence included in the explanation - with the intent to rapidly produce a concise explanation, and not necessarily the most complete one, our main intention was to investigate whether our produced concise explanation does not miss factors that clinicians considered significant and not whether extra variables are included in the explanation.

We defined 4 groups based on the percentage of the variables that were considered as significant by clinicians and were also part of the provided explanation. Thus, their qualitative answers were categorised into the following groups:

1. Not at all similar (clinicians' reasoning is 0% similar to the explanation)
2. Quite similar (clinicians' reasoning is 1 - 49% similar to the explanation)
3. Similar (clinicians' reasoning is 50 - 74% similar to the explanation)
4. Very similar (clinicians' reasoning is ≥75% similar with the explanation)

Suppose that for a specific case, clinicians mentioned as significant the variables long bone fracture, unstable pelvis, FAST scan, Lactate and the first level of the explanation has mentioned as significant the variables unstable pelvis, Lactate, and GCS. As two out of the four variables mentioned by clinicians were also part of the explanation, the similarity belonged to group 3. As we focused on not missing any variable mentioned by the clinicians, negative weight was not given in cases where the explanation included more variables than those mentioned by clinicians. The aim was to compare how similar their reasoning was to our explanation based on the same available evidence. As a result, only clinicians' reasoning in the baseline was examined.

Clinicians' *trust* in the model's prediction between the *prediction* and the *explanation* cluster was compared, using their answers to the seven-point scale question: 'How much would you say that you trust the prediction of the model?'.

The analysis of the *potential benefit* was based on three questions. First, we compared their assessment of coagulopathy not only in the baseline and the follow-up but also in the *prediction* and the *explanation* cluster. Then, based only on the follow-up questionnaires, we examined how useful the model was. The usefulness was two-fold: (i) confirmation of their assessment and (ii) revision of their assessment. Finally, we compared their baseline and follow-up answers to the question 'What would you do next?' to examine whether the extra information had an impact on their decision-making process.

The *clarity* of the explanation was based on the final question of the *explanation* cluster 'How clear was the explanation of the prediction of coagulopathy?' and clinicians' feedback.

## 6.4 Results

We present the results in each of the 4 categories: *similarity*, *trust*, *potential benefit* and *clarity*.

### 6.4.1 Similarity

To assess the *similarity*, we wanted to examine whether at least half of clinicians' responses are similar to our explanation. Having 160 responses (10 clinicians reviewed independently 16 cases) and classifying as similar those responses that refer to categories 3 and 4, we found that 71% of clinicians' responses were similar to our explanation. Based on a one-tail proportion test we had enough evidence to reject the null hypothesis and support our claim that clinicians' reasoning was ≥50% similar to our explanation (*p-value < 0.001*).

### 6.4.2 Trust

We wanted to examine whether clinicians' *trust* on the model's prediction was significantly greater in the *explanation* than the *prediction* cluster. Each case was seen by a matched pair of clinicians (80 responses); one clinician viewed only the model's prediction (*prediction* cluster) and the other was also exposed to an explanation of the model's reasoning (*explanation* cluster). Using the Wilcoxon matched-pairs signed-ranks test we did not have enough evidence to reject the null hypothesis (*p-value = 0.82*).

### 6.4.3 Potential Benefit

The first element of the *potential benefit* that we investigated was whether there was any significant difference in their assessment of coagulopathy between the baseline and the follow up questionnaire. Wilcoxon matched-pairs signed-ranks test showed that we had enough evidence to reject the null hypothesis (*p-value = 0.003*) and support our claim.

In addition, we examined whether, for the coagulopathic patients, clinicians' assessment in the *explanation* cluster was greater than their assessment in the *prediction* cluster. Using the Wilcoxon matched-pairs signed-ranks test we had enough evidence to reject the null hypothesis (*p-value = 0.048*) and support our claim. Similarly, we tested whether clinicians' assessment for the non-coagulopathic patients was lower in the *explanation* cluster than the *prediction* cluster. We did not have enough evidence to reject the null hypothesis (*p-value = 0.98*).

Clinicians answered a seven-point scale question: 'How useful was the prediction of coagulopathy for confirming your assessment?'. We examined whether their rate of the model's usefulness for confirming their assessment was greater in the *explanation* than the *prediction* cluster. Using the Wilcoxon matched-pairs signed-ranks test we did not have enough evidence to reject the null hypothesis (*p-value=0.36*). The same analysis was conducted about the usefulness of the model in revising their assessment. Again, not enough evidence was available for rejecting the null hypothesis (*p-value=0.81*).

Finally, the extra information had no impact on their decision-making process. In the baseline and the follow-up assessment their actions were the same.

### 6.4.4 Clarity

We wanted to examine whether at least half of the provided explanation was considered clear. Having 80 responses (we used only the *explanation* cluster) and classifying as clear the explanations that were rated with 5,6 or 7, we found that 65.2% of the provided explanations were classified as clear by clinicians. Based on a one-tail proportion test we had enough evidence to reject the null hypothesis and support our claim (*p-value = 0.01267*). Finally, in the *explanation* cluster clinicians gave their feedback on the explanation in an open question. Some of their comments were:
- *In the heat of battle, a colour coded guidance would aid clarity*
- *The words 'partially support' and 'partially do not support' are not very clear*
- *Why haemothorax (HT) at the beginning was very important and at level 3 was partially supporting*
- *Weighting leans towards more significant chance of coagulopathy*
- *Lactate 4.5 is non-supportive, 4.6 would be supportive?*
- *Level 1: useful brief explanation, level 3: too wordy*
- *ATC reassuring when agrees with my prediction*
- *Expected higher prediction*

## 7 Discussion

As BNs become more complex, the difficulty of understanding the model's reasoning grows too. Decision models are built to improve decision making. Consequently, an explanation of how a model came to a prediction is an important part of model's use and trust. In this paper we have described an algorithm that generates an explanation in three levels, each adding more details to the explanation. Our method can be applied to BNs with both discrete and continuous variables and requires no user input. It is suitable for real-time use, as it focuses on rapidly producing a good explanation and not necessarily the most complete one. Our algorithm produces quick and concise explanations, but we recognize some limitations.

One major limitation of our algorithm is the way we classify the evidence as influential. There are two quick approaches to identify the set of influential evidence: (i) observe each evidence variable in turn and compare the posterior probability with the prior (ii) remove each evidence variable in turn and compare the posterior with the posterior with all the evidence. Each approach focuses on a different type of influence, so each can miss detecting some influences under certain circumstances.

Imagine that we have a binary target *T*: {*True, False*} and two evidence variables *A*: {*True, False*} and *B*: {*True, False*}, as shown in Figure 11. Two scenarios are available; (1) the set of evidence is *E*: {*A = True, B=True*} and (2) the set of evidence is *E*: {*A = False, B=True*}. Suppose that in both scenarios our aim is to identify whether the evidence variable *B* is influential. Using approach (i) the impact of *B* is identified by comparing the posterior when B is instantiated *P(T|B)* with the prior *P(T)*. Using the Hellinger distance $D_H(P(T|B)||P(T))$, the impact of *B* is 0.14 in both scenarios. This approach cannot distinguish between the two different scenarios, as it neglects the complete set of evidence. On the contrary, using approach (ii) the impact of *B* is identified by comparing the posterior with all the evidence *P(T|E)* with the marginal posterior when *B* is temporally removed *P(T|E\B)*. Using the Hellinger distance $D_H(P(T|E)||P(T|E\backslash B))$, the impact of *B* is 0.11 and 0.23, in scenarios (1) and (2), respectively. Now that the whole set of evidence is considered, the impact of *B* is different in each scenario.

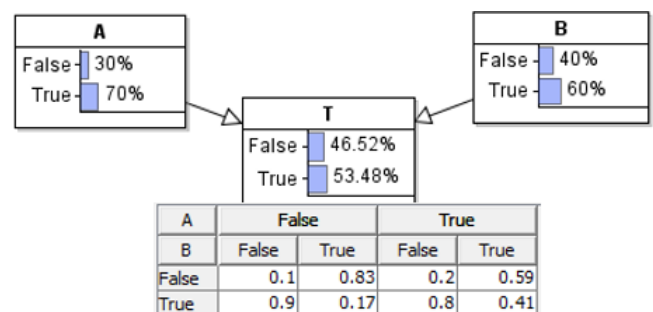

Figure 11. A three-node BN

Removing each evidence separately can distinguish different scenarios. However, there are situations, such as having AND or OR operator or mutually exclusive causes, where some influential evidence may be missed using approach (ii). For instance, imagine that *T=True* if both **A** and **B** are true (AND operator). In case **E**: {*A = False, B=True*}, using approach (ii) the impact of **B** is $D_H(P(T|E)||P(T|E\backslash B)) = 0$. Therefore, when we have the AND operator and at least one parent is false, using approach (ii) we may miss some of the influential evidence. Similarly, when we have the OR operator and at least one parent is true. This problem becomes more important when we have more evidence. On the contrary, approach (i) gives a non-zero influence in both situations.

A combination of the two approaches would be more appropriate under certain conditions. However, we chose to use approach (ii) and remove each evidence, as it is crucial to account for the rest of the available evidence. This is closer to the real world and the way people think about explanation. We removed one evidence at a time and not combinations of the evidence to reduce the time to produce the explanation. We accept the risk of missing interactions since the time needed to search for the best combination of evidence increases exponentially when the model and the number of evidence variables become bigger. In addition, for producing a concise and sufficient explanation, and not necessarily the most complete one, which is applicable in situations where there is a time pressure, no user input was required for selecting the set of significant evidence. This is a necessary restriction when time is of an essence, but we acknowledge that it makes the procedure inflexible.

The MB of the target is used as the intermediate step to capture the flow of information from the evidence to the target. The advantages of using the MB are: (1) every variable in a BN has a MB, so it can be generalised, (2) the MB of a variable contains important information about it, and (3) it can be used to produce a meaningful explanation very quickly. However, in large BNs, where the evidence is further from the target, important variables along the chain will not be captured in the explanation. We could overcome this limitation by adding another explanation level, if needed, in which we present the MB of the MB variables.

We also recognize the limitations of the evaluation study. Although based on a sound design, this study was based only on 10 real cases, so it cannot give definitive conclusions. It was used as an initial pilot study to investigate the potential benefits and shortcomings of the explanation, and to teach us useful lessons for a future larger trial. This study primarily looked at the similarity between clinicians' reasoning and the generated explanation and at the increase in model's trustworthiness. Secondarily, we examined the potential benefit of the explanation on clinicians' decision making and assessment. Finally, the clarity of the explanation was tested.

The explanation produced by our method was able to identify in most of the cases the majority of the evidence that clinicians mentioned as significant. As our aim was to assess whether our concise explanation did not miss the significant evidence identified by the clinicians, extra evidence that were part of the provided explanation was not considered as discrepancies between the explanation and clinicians reasoning. Clinicians trusted the model's prediction, but there was no significant change in their trust when an explanation was provided. There was no impact on their decision making but there was a significant change in their assessment. However, there was not enough evidence to support the usefulness of the provided explanation in conforming or revising their assessment. The explanation was found to be clear but very wordy by most clinicians. They liked the first level of the explanation but found the third level of the explanation too complicated. They liked the idea of having an explanation but would prefer it to be less wordy and potentially graphically enhanced.

Another limitation of the study was the chosen cases. Although they had a degree of ambiguity as coagulopathy is an uncertain condition, they were almost always similar with clinicians' expectations. That might explain why the explanation did not have a significant impact on the model's trust and clinicians' decision making. Coagulopathy is a disease that takes time to develop. When a trauma patient arrives in the ED there are some standard actions that clinicians could carry out, such as examine the patient, give blood, go to theatre, conduct extra imaging etc. Having a justified prediction of coagulopathy can reassure their beliefs but it is not going to make them change their decisions. This was potentially the reason why the explanation did not have a significant impact on their decision making. In addition, the chosen clinicians were very experienced, so a decision tool and an explanation may have less significant benefit on their decision making than it would have on inexperienced clinicians. This can also justify the fact that they answered that the explanation was not very useful for confirming or revising their assessment, even if their assessment was significantly improved, especially for the coagulopathic patients. Finally, the length of the explanation could be an inhibiting factor.

## 8   Conclusion

We proposed a quick way to generate a concise explanation of reasoning for BNs that contain both discrete and continuous nodes, without any further input from the user. The small evaluation study showed that the explanation is meaningful and similar to clinicians reasoning. It does not have a significant impact on the model's trust and clinicians' decision making but it can affect clinicians' assessment. Clinicians liked having an explanation, while they preferred it to be simpler and less word laden.

To reduce the time to search for the best combination of evidence, only one item of evidence was removed at a time. This restriction may miss detecting some important evidence under certain circumstances. In addition, no user input was required for selecting the set of significant evidence, resulting in an explanation that is applicable in time critical situations but also in an inflexible approach. Furthermore, for saving time we only used the MB of the target as the intermediate step in the reasoning process. This can help us to generate a meaningful explanation quickly, but important information, especially in large BNs, might be missed. A useful next step is to investigate how we can prune the available evidence and intermediate variables, using the knowledge of the model's structure and the domain knowledge. For instance, it would be useful to generate an explanation by making use of abstract semantics, such as idioms [32], [33]. In addition, the proposed method targets only BNs that are used as a one-time activity. A useful extension would be to investigate how our explanation can be extended to time-based BNs, where the explanation generated for a target in a later stage should distinguish between evidence entered in the same and in earlier stages. An investigation of the produced explanation in situations where multiple target variables are available would be informative. An enhanced graphical representation and an evaluation of the explanation in real time would help us to examine how much a decision maker makes use of the explanation under real conditions and potentially time pressure.

**Acknowledgement**

This work was supported in part by European Research Council Advanced Grant ERC-2013-AdG339182-BAYES KNOWLEDGE. The authors EK and WM acknowledge support from the EPSRC under project EP/P009964/1: PAMBAYESIAN: Patient Managed decision-support using Bayes Networks. The author SM acknowledges support from Dept of Research & Clinical Innovation, HQ Joint Medical Grp, UK Defence Medical Services.

# References

[1]   P. J. F. Lucas and A. Abu-Hanna, "Prognostic methods in medicine," *Artif. Intell. Med.*, vol. 15, pp. 105–119, 1999.

[2]   J. C. Wyatt and D. G. Altman, "Commentary: Prognostic models: clinically useful or quickly forgotten?," *BMJ*, vol. 311, pp. 1539–1541, Dec. 1995.

[3]   P. Lucas, "Bayesian networks in medicine: a Model-based Approach to Medical Decision Making," in *EUNITE workshop on Intelligent Systems in Patient Care*, 2001.

[4]   D. B. Toll, K. J. M. Janssen, Y. Vergouwe, and K. G. M. Moons, "Validation, updating and impact of clinical prediction rules: a review.," *J. Clin. Epidemiol.*, vol. 61, pp. 1085–1094, Nov. 2008.

[5]   K. G. M. Moons, D. G. Altman, Y. Vergouwe, and P. Royston, "Prognosis and prognostic research: application and impact of prognostic models in clinical practice.," *BMJ*, vol. 338, no. jun04 2, p. b606, 2009.

[6]   C. Lacave and F. J. Díez, "A review of explanation methods for Bayesian networks," *Knowl. Eng. Rev.*, vol. 17, no. 2, pp. 107–127, Apr. 2002.

[7]   C. Lacave and F. J. Díez, "A review of explanation methods for heuristic expert systems," *Knowl. Eng. Rev.*, vol. 19, pp. 133–146, 2004.

[8]   K. O. Pedersen, "Explanation Methods in Clinical Decision Support," 2010.

[9]   A. Holzinger, G. Langs, H. Denk, K. Zatloukal, and H. Müller, "Causability and explainability of artificial intelligence in medicine," *Wiley Interdiscip. Rev. Data Min. Knowl. Discov.*, vol. 9, no. 4, pp. 1–13, 2019.

[10]  E. Kyrimi and W. Marsh, "A Progressive Explanation of Inference in 'Hybrid' Bayesian Networks for Supporting Clinical Decision Making," in *the Eighth International Conference on Probabilistic Graphical Models*, 2016, vol. 52, pp. 275–286.

[11]  J. Pearl, *Probabilistic reasoning in intelligent systems: networks of plausible inference*. Morgan Kaufmann Publishers Inc., 1988.

[12]  G. Harman, "The Inference to the Best Explanation," in *The Philosophical Review*, vol.

74, no. 1, 1965, pp. 88–95.

[13] P. Lipton, "Inference to the Best Explanation. Second edition," in *London: Routledge*, 2004.

[14] J. Hintikka, "What is Abduction? The Fundamental Problem of Contemporary Epistemology," *Trans. Charles S. Peirce Soc.*, vol. 34, no. 3, p. 503, 1998.

[15] G. Minnameier, "Peirce-suit of truth – why Inference to the Best Explanation and abduction ought not to be confused," *Erkenntnis*, vol. 60, no. 1, pp. 75–105, 2004.

[16] D. G. Campos, "On the distinction between Peirce's abduction and Lipton's Inference to the best explanation," *Synthese*, vol. 180, no. 3, pp. 419–442, 2011.

[17] B. Yet, Z. B. Perkins, N. R. M. Tai, and D. W. R. Marsh, "Clinical evidence framework for Bayesian networks," *Knowl. Inf. Syst.*, 2016.

[18] S. T. Timmer, J.-J. C. Meyer, H. Prakken, S. Renooij, and B. Verheij, "Explaining Legal Bayesian Networks Using Support Graphs," in *Legal Knowledge and Information Systems. JURIX 2015: The Twenty-eighth Annual Conference*, 2015, pp. 121–130.

[19] C. S. Vlek, H. Prakken, S. Renooij, and B. Verheij, "A method for explaining Bayesian networks for legal evidence with scenarios," *Artif. Intell. Law*, vol. 22, no. 4, pp. 375–421, 2016.

[20] Q. Zhong, X. Fan, X. Luo, and F. Toni, "An explainable multi-attribute decision model based on argumentation," *Expert Syst. Appl.*, vol. 117, pp. 42–61, 2019.

[21] S. T. Timmer, J. J. C. Meyer, H. Prakken, S. Renooij, and B. Verheij, "A two-phase method for extracting explanatory arguments from Bayesian networks," *Int. J. Approx. Reason.*, vol. 80, pp. 475–494, 2017.

[22] H. J. Suermondt, "Explanation in bayesian belief networks," *PhD thesis, Stanford Univ.*, 1992.

[23] U. Chajewska and J. Halpern, "Defining Explanation in Probabilistic Systems," *Proc. Thirteen. Conf. Uncertain. Artif. Intell.*, pp. 62–71, 1997.

[24] P. Haddawy, J. Jacobson, and C. E. Kahn Jr., "BANTER: a Bayesian network tutoring shell," *Artif. Intell. Med.*, vol. 10, pp. 177–200, 1997.

[25] I. J. Good, "Explicativity: a mathematical theory of explanation with statistical applications," in *Royal Statistical Society of London*, 1977, pp. 303–330.

[26] D. Madigan, K. Mosurski, and R. G. Almond, "Graphical explanation in belief networks," *J. Comput. Graph. Stat.*, vol. 6, no. 2, pp. 160–181, 1997.

[27] P. Sutovský and G. F. Cooper, "Hierarchical explanation of inference in Bayesian networks that represent a population of independent agents," in *18th European Conference on Artificial Intelligence*, 2008, pp. 214–218.

[28] J. Leersum, "Explaining the reasoning of Bayesian networks," *Master Thesis, Utr. Univ.*, 2015.

[29] M. Neil, M. Tailor, and D. Marquez, "Inference in Hybrid Bayesian Networks using

Dynamic Discretisation," *Stat. Comput.*, vol. 17, no. 3, pp. 219–233, 2007.

[30] Agena Ltd, "AgenaRisk." 2018.

[31] B. Yet, Z. Perkins, N. Fenton, N. Tai, and W. Marsh, "Not just data: a method for improving prediction with knowledge.," *J. Biomed. Inform.*, vol. 48, pp. 28–37, Apr. 2014.

[32] N. Martin, N. Fenton, and L. Nielsen, "Building large-scale Bayesian networks," *Knowl. Eng. Rev.*, vol. 15, no. 3, pp. 257–284, 2000.

[33] E. Kyrimi, M. Neves, M. Neil, S. McLachlan, W. Marsh, and N. Fenton, "Medical idioms for clinical Bayesian network development," *Manuscr. Submitt. Publ.*, 2019.